\documentclass[letterpaper]{article} 
\usepackage{aaai24}  
\usepackage{times}  
\usepackage{helvet}  
\usepackage{courier}  
\usepackage[hyphens]{url}  
\usepackage{graphicx} 
\usepackage{natbib}  
\usepackage{caption} 
\usepackage{xcolor}
\usepackage[ruled,lined]{algorithm2e}
\DontPrintSemicolon

\usepackage{booktabs}
\usepackage{multirow}
\usepackage{makecell}
\usepackage{rotating}

\usepackage{amsmath}

\definecolor{periwinkle}{HTML}{7977B8}

\newcommand{\bo}[1]{\textbf{#1}}
\newcommand{\bw}[1]{\textbf{\textcolor{periwinkle}{#1}}}

\usepackage{amsthm}

\makeatletter
\renewcommand{\SetKwFunction}[2]{%
  \expandafter\gdef\csname @#1\endcsname##1{\FuncSty{#2\big(}\FuncArgSty{##1}\FuncSty{\big)}}%
  \expandafter\gdef\csname#1\endcsname{%
    \@ifnextchar\bgroup{\csname @#1\endcsname}{\FuncSty{#2}\xspace}}%
}%
\makeatother

\usepackage{mathtools}
\DeclarePairedDelimiter\ceil{\lceil}{\rceil}

\usepackage[shortlabels]{enumitem}

\title{Fast Feedforward Networks}
\author{
    Peter Belcak and Roger Wattenhofer
}
\affiliations{
    ETH Zürich \\
    \{belcak,wattenhofer\}@ethz.ch
}

\begin{document}

\maketitle

\begin{abstract}
We break the linear link between the layer size and its inference cost by introducing the fast feedforward\footnote{\texttt{https://github.com/pbelcak/fastfeedforward}\label{footnote:code_link}} (FFF) architecture, a $\log$-time alternative to feedforward networks.

We demonstrate that FFFs are up to 220x faster than feedforward networks, up to 6x faster than mixture-of-experts networks, and exhibit better training properties than mixtures of experts thanks to noiseless conditional execution.

Pushing FFFs to the limit, we show that they can use as little as 1\% of layer neurons for inference in vision transformers while preserving 94.2\% of predictive performance.
\end{abstract}

\section{Introduction}
\label{sec:introduction}

The feedforward layer is a parameter-heavy building block of transformer models \citep{vaswani2017attention}.
Growing to tens of thousands of hidden neurons in recent years, the cost of feedforward layer inference is now in the sights of those seeking to make large models faster.

It has been recognized that in very large networks, only a small portion of the feedforward hidden neurons plays a role in determining the output for any single input, and that it is possible to design networks that are modular in order to utilize this fact \cite{bengio2015conditional}.

The most recent work on the modularization of feedforward layers aims at architectural designs that implicitly encourage sparsity \cite{shazeer2017outrageously,lepikhin2020gshard,fedus2022switch}.
They share the common approach of subdividing the feedforward layer into separate blocks of neurons -- ``experts'' -- and training a gating layer to determine the mixture of experts to be used in the forward pass.
Inference acceleration is then achieved by using only the best-scoring $k$ blocks, or a variant thereof.
This approach scales down the inference time by a constant but remains linear in the width of the feedforward layer.
Moreover, it relies on noisy gating to allow for load balancing among the experts, complicating training and encouraging duplicity.

\begin{figure}[t!]
    \centering
    \includegraphics[width=\linewidth]{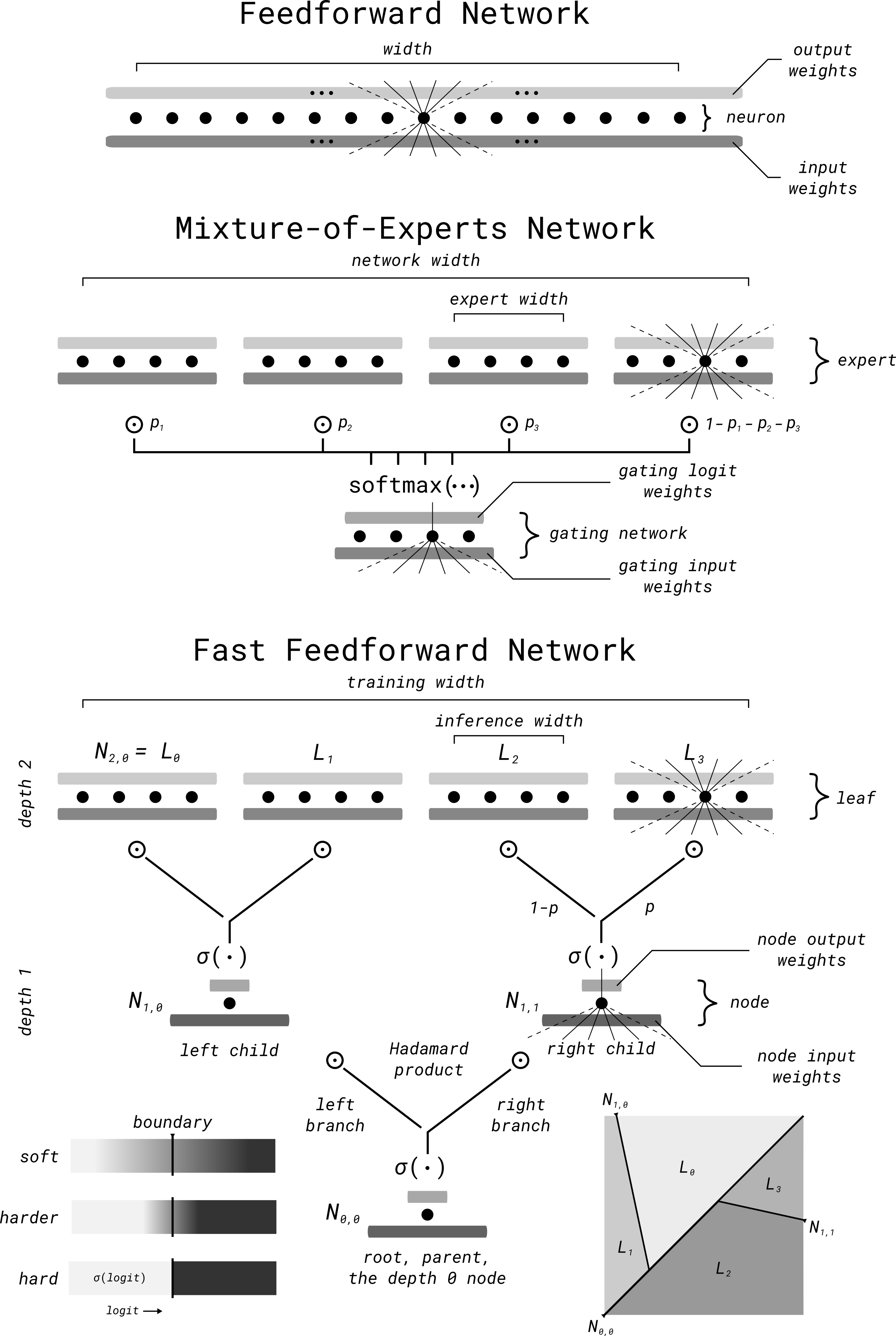}
    \caption{
        A fast feedforward network set in comparison to its peers.
        \textit{Bottom.} Illustrations of the resulting regionalization of the input space and varying boundary hardness.
    }
    \label{figure:sketch}
    \vspace{-12pt}
\end{figure}

\paragraph{Outline.} We introduce the \textit{Fast Feedforward (FFF)} architecture -- a peer of the feedforward (FF) architecture that accesses blocks of its neurons in logarithmic time.
FFF divides the input space into disjoint regions by means of a differentiable binary tree and performs simultaneous learning of the region boundaries and the neural blocks assigned to these regions.
This is achieved by tree-conditional execution of neurons: a small subset of \textit{node} neurons is set apart to choose what mixtures of  \textit{leaf} neuron blocks are to be computed to produce the final output (Figure~\ref{figure:sketch}).
As the training progresses, the region boundaries harden, and the mixtures tend toward selecting only one $\log$-time-accessible leaf.

\noindent
Formally, let $f: \mathcal{D}_I \to \mathcal{D}_O$ be the learning target.

The naive approach is to train a feedforward layer (FF) $F$ of width $w$ to approximate $f$ on $\mathcal{D}_I$, i.e. $F \approx_{\mathcal{D}_I} f$.

The mixture-of-experts (MoE) approach \cite{shazeer2017outrageously} is to choose an expert width $e$ that does not hinder performance, and then train separate expert blocks $E_1,\dots,E_{\ceil{w/e}}$ of neurons mixed by the partially randomized output of a gating network of width $g=\ceil{w/e}$.
The learning target $f$ is then approximated on $\mathcal{D}_I$ under the mixture of $k$ best-scoring experts, i.e. $m_1E_{b_1} + ... + m_kE_{b_k} \approx_{\mathcal{D}_I} f$.
For the corresponding FF network of width $w$, a MoE network with $g=\ceil{w/e}$ experts uses $ke$ neurons for inference at the mixture overhead of $g$.

Fast feedforward networks (FFFs) are designed to leverage the fact that different regions of the input space activate different sets of neurons in wide networks.
FFFs of depth $d$ jointly learn a tree partition $R_1,\dots,R_{2^d}$ of the input space determined by their nodes, and $2^d$ small leaf feedforward networks $L_1,\dots,L_{2^d}$ of width $\ell$, which are trained so that each $L_i$ approximates the partial target function $f|_{R_i}$ on $R_i$, i.e. $L_i \approx_{R_i} f$.
Crucially, an FFF with depth $d$ and leaf width $\ell$ can use all $2^d\ell$ hidden neurons to learn $f$, but requires only one leaf of width $\ell$ to compute its output for any $\iota \in \mathcal{D}_I$, and does so at the lookup overhead of only $d$ neurons.

To draw a comparison, the leaves $L_i$ are to FFFs what experts $E_j$ are to MoEs, but the FFF tree network is regionalizing the input space rather than noisily voting on expert prowess.
For the corresponding feedforward network of width $w$ and a choice of leaf size $\ell$, one can choose $d = \log_2 \ceil{w/\ell}$ and access the leaf in $\mathcal{O}(d) = \mathcal{O}(\log{w})$ instead of $\mathcal{O}(g) = \mathcal{O}(w)$ time.
Coincidentally, the FFF approach also happens to represent a differentiable relaxation of the classical notion of $k$-d trees \cite{bentley1975multidimensional}.

The process enabling the fragmentation of a large feedforward layer to a number of smaller leaf layers while preserving its predictive performance is \textit{hardening}.
In training, the nodes of FFF recursively perform a soft choice $\left<1-p, p\right>$ over the outputs of their two children (a ``mixture of child experts'' if we must), and based on the final loss incurred, the optimizer updates both the weights of the children and the weights of the parent that computed the choice.
During inference, a single hard decision (i.e. ``proceed to left/right child'') is made depending on the rounding result $[p]$ at each node.
Hardening is the process of nodes learning boundaries such that the soft choices they make for individual inputs go from indecisive mixtures (e.g. $\left<49, 51\right>$) toward more decisive ones (e.g. $\left<3,97\right>$).
We observe that in settings where representation power is plentiful (i.e. wide leaves and deep FFFs), the process often takes place on its own.
In settings where more representational power may be warranted (e.g. wide vision transformers with FFF leaf bottlenecks), this process either takes place but stalls prematurely, or takes place at a very low rate, and we choose to encourage it with the addition of a \textit{hardening loss}.

Thanks to hardening, the performance of the soft training setting carries over to inference.
As a byproduct, the learned regions can also be used as a partition of the input space for interpretability, surgical model editing, catastrophic forgetting mitigation, reduction of replay data budget, etc..

\paragraph{User manual.}
Suppose that you have an existing architecture featuring a FF layer of width $w$ and want to replace it with a FFF layer.

\textit{Case 1: ``I want faster inference''.}
Choose $\ell << w$ such that $\ell$ fits your target inference budget, and then experiment with different depths $d \geq \log_2 (w / \ell)$ to achieve the desired performance.
Note that $\ell$ still needs to be large enough to be able to learn the partial function on its region, and that the final training width $2^d\ell$ might end up being larger than $w$.

\textit{Case 2: ``I want a partition of the input space''.}
Choose $d$ such that $2^d$ meets your expectation on the number of regions.
Then experiment with $\ell \geq w2^{-d}$ for best performance.
Note that $\ell$ again needs to be large enough to be able to learn the partial function on its region, and that for large $d$ you might have to actively counter the effects of overfragementation.

\paragraph{Contributions.}

\begin{enumerate}
    \item We introduce the fast feedforward (FFF) architecture, a peer to the feedforward (FF) architecture that uses only a log-time-accessible fraction of its neurons at any point.

    \item We investigate the effect of leaf size and depth on the predictive performance of FFFs as models in their own right and show that in sufficiently large settings, FFFs give performance comparable to FFs of the same training width while carrying out the inference significantly faster.
    We further show that FFFs deliver better memorization and generalization performance than the FFs of the same inference size.

    \item We compare the FFF architecture against the mixture-of-experts (MoE) approach in terms of their predictive performance and inference speed as the number of blocks increases, and support the claimed advantages of the design experimentally.

    \item We demonstrate that FFFs can be feasibly used in place of FFs as parts of larger, more complex architectures such as transformers.
\end{enumerate}
The text culminates into a comparison with related work.

\section{Algorithm}
\label{sec:algorithm}
Denote the network input, output dimension $\dim_I, \dim_O$.
Denote $\left<a,b,c\right>$-feedforward network a feedforward network with $a$ inputs, $b$ neurons, and $c$ outputs.

Notice that we override the terminology designed for multi-layer networks and talk of only one set of neurons that has both input and output weights.
For example, we would refer to the BERT-base feedforward layer with input dimension 768, 3072 hidden neurons, and 768 output neurons as to the feedforward layer with 3072 neurons, each with 768 inputs and 768 outputs.
This greatly simplifies our presentation.

\paragraph{Definition.} A fast feedforward network of depth $d \geq 0$, node size $n$, and leaf size $\ell$ is a pair $\left<\mathcal{N}, \mathcal{L}\right>$.

$\mathcal{N} := \{N_{0,0},\dots,N_{d-1,2^{d-1}-1}\}$ is the set of node $\left<\dim_I,n,1\right>$-feedforward networks with an additional sigmoidal activation on the output. These nodes form a balanced differentiable binary tree such that $N_{m+1,2n},N_{m+1,2n+1}$ are the children of $N_{m,n}$.

$\mathcal{L} := \{N_{d,0},\dots,N_{d, 2^{d}-1}\}$ is the set of leaf $\left<\dim_I,\ell,\dim_O\right>$-feedforward networks.
All weights are trainable by default and the forward pass is governed by the fully deterministic Algorithm \ref{algorithm:forward_pass}.

\begin{algorithm}[t]
  \KwIn{Input sample $\iota \in \mathcal{D}_I$, the root node $N_{0,0}$}
  \KwOut{Output $\in \mathcal{D}_O$ of the FFF for $\iota$}
  \SetKwFunction{FT}{$\textsc{Forward}_T$}
  \SetKwProg{Fn}{Function}{:}{}
  \;
  \Fn{\FT{$\iota$, $N_{m,n}$}}{
        \uIf{$N_{m,n} \in \mathcal{L}$}{
            \KwRet{ $N_{m,n}\left(\iota\right)$ }\;
        }
        \Else{
            $c_{m,n} \gets N_{m,n}\left(\iota\right)$\;
            \KwRet{ $c_{m,n}\,$\FT{$\iota, N_{m+1,2n+1}$} $\, +\, (1-c_{m,n})$\,\FT{$\iota, N_{m+1,2n}$} }\;
        }
  }
  \;
  \SetKwFunction{FI}{$\textsc{Forward}_I$}
  \SetKwProg{Fn}{Function}{:}{}
  \Fn{\FI{$\iota$, $N_{m,n}$}}{
        \uIf{$N_{m,n} \in \mathcal{L}$}{
            \KwRet{ $N_{m,n}\left(\iota\right)$ }\;
        }
        \Else{
            $c_{m,n} \gets N_{m,n}\left(\iota\right)$\;
             \uIf{$c_{m,n} \geq \frac{1}{2}$}{
                \KwRet{ \FI{$\iota, N_{m+1,2n+1}$} }\;
            }
            \Else{
                \KwRet{ \FI{$\iota, N_{m+1,2n}$} }\;
            }
        }
  }
  \caption{FFF forward pass.}
  \label{algorithm:forward_pass}
\end{algorithm}

The nodes of FFF are arranged in a differentiable $\dim_I$-d tree that makes a soft choice over the leaves in the form of a stochastic vector $c$.
In training, FFF performs a mixture of experts over all leaves in $\mathcal{L}$, with the choice weights of the mixture $c$ computed by ascending through the tree from the root node $N_{0,0}$ (cf. $\textsc{Forward}_T$).
During inference, the decision at each node is taken to be the closer of $\left\{0, 1\right\}$, and the forward pass algorithm proceeds from the root, always choosing only one branch depending on the local node decision (cf. $\textsc{Forward}_I$).

\paragraph{Regions of responsibility and their boundaries.}
The tree component of the FFF yields a partition of the input space.
Each leaf is responsible for exactly one region of this partition, even though during training, its prediction on its own region of responsibility is mixed with the predictions of other leaves.
The boundaries between the individual regions are determined by the node networks.

In the case when $n=1$ and there is no activation on the node network but the head sigmoid, the boundary is the activation plane of the hidden neuron.
The norm of the plane's normal vector (=weights of the neuron) affects how quickly the sigmoid goes from $0$ to $1$ around the boundary (cf. Figure~\ref{figure:sketch} bottom-left).
This determines how clearly the boundary is defined. 

\paragraph{Hardening.}
For the predictive performance of $\textsc{Forward}_T$ to carry over to $\textsc{Forward}_I$, one must not lose predictive information when rounding the choice scores.
This loss of information is minimal when the boundary decisions have been properly hardened (cf. Introduction).
As hinted at above, hardening at a node generally does not have to involve adjustment to the boundary as a manifold in space -- progressive uniform rescaling of boundary coefficients (i.e. squashing of the final sigmoid toward the step function to make the boundary more clearly defined) suffices.

Interpreting the node choice scores as Bernoulli probabilities, hardening can be tracked by monitoring the batch mean of entropies of the choices scores at each node.
In our experimentation, we found that rounding choice pairs $\left<1-p,p\right>$ with entropies below $0.10$ tends to lead to only very modest deviations from the $\textsc{Forward}_T$ performance.
For situations where the hardening of node decisions does not occur to a sufficient extent on its own, hardening can be encouraged by the addition of \textit{hardening loss}.

Let $L_{\text{pred}}$ be the loss due to the outputs of FFF.
Then one can take the total loss to be $L_{\text{total}} := L_{\text{pred}} + h L_{\text{harden}}$ with
\[
    L_{\text{harden}} := \sum\limits_{\iota \in \mathcal{B}} \sum\limits_{N \in \mathcal{N}} H\left(N(\iota)\right),
\]
where $\mathcal{B} \subseteq \mathcal{D}_I$ is a batch of samples, $H(p)$ the entropy of a Bernoulli random variable, and $h$ the training hyperparameter controlling the effect of the hardening loss.

\paragraph{Overfragmentation.}
If pushed to the extreme, allowing fast feedforward networks to learn too many hard boundaries leads to \textit{overfragmentation} -- a phenomenon in which the network divides the input space into exponentially many disjoint regions and learns to approximate parts of the learning target in a way that is too specific for each region.
Overfragmentation has two direct consequences: localized overfitting and the ``shrinking batch problem''. 

\textit{Localized overfitting} denotes a tail process that occurs once the model has sufficiently hardened, in which the region boundaries are no longer flexible and certain leaves learn to overfit the training data on their regions of responsibility.
This is because they stop receiving meaningfully large gradient updates from the neighboring regions, but may be responsible for handling test samples that are not well understood by the training data for their region.
Localized overfitting manifests itself just like classical overfitting -- the validation performance ceases to improve or deteriorates while learning on the training set continues.
It can be mitigated by randomized child transpositions -- the soft decisions $\left<1-p,p\right>$ at each node can be randomly transposed with some low probability into $\left<p,1-p\right>$.
This is to expose their children to the training data of neighboring regions in order to aid generalization performance and does to some extent already happen for soft boundaries, but it becomes rare as the boundaries harden.

The FFF variant of \textit{the shrinking batch problem} is also a result of the leaf region boundaries hardening, and it arises in situations when the batch size becomes too small for the partition of the input space learned by the FFF tree.
If the partition of the input space is too finely grained and the boundaries hardened, each leaf ends up receiving meaningful gradient updates from only a small fraction of the training samples, resulting in inaccurate gradient descent progress. 
Batch shrinking leads to poor learning performance (e.g. low training set accuracy, early stalling, chaotic development) but can be mitigated -- naively with larger batch sizes, gradient accumulation, and smaller learning rates; in full generality with localized optimization. 

\smallskip
\noindent
We consider the complexity of our algorithms in terms of the parameters $d,n,\ell$. Note that we found $n=1$ to suffice in all our experiments, but we keep $n$ in for the sake of generality.

\paragraph{Training complexity.}
The training forward pass $\textsc{Forward}_T$ ascends through $d$ levels of the tree, passing through node neurons to compute the final choice vector $c$ in $\mathcal{O}((2^d - 1)n)$ time.
Then, the leaf outputs are computed and mixed by $c$ in $\mathcal{O}(2^d \ell)$ time.
This means  $\mathcal{O}(2^d (\ell + n) - n)$ time for the forward pass, and a $(d+1)$-step backward pass back to the decision on the root.

From the implementation standpoint, we express the ascent through the tree as a single loop making $d$ identical batched computations, and then perform the final leaf forward pass and expert mixture.

\paragraph{Inference complexity.}
$\textsc{Forward}_I$ ascends through the FFF tree in $d$ steps, always executing exactly one node network.
Then, it performs inference on one leaf, leading to $\mathcal{O}(dn + \ell)$ time.

In terms of the implementation, the ascent from the root through the tree is executed as a batched computation of an indexed set of weights and biases (multiply-and-accumulate), comparison of the logit to $0$, and advancing of the index depending on the result of the comparison.

In our experience of using ahead-of-time compilation for CUDA, the selective indexing of weights for node decisions manifested itself in the native code as a simple offset in the data load for batched matrix multiplication, having only a small constant implementation overhead on the hardware level when compared to feedforward layers.

\paragraph{Size and width.}
Fast feedfoward networks consist of neurons of two types: node and leaf neurons.
For clarity and to make direct comparisons to the corresponding feedforward networks, we distinguish between variants of network size and width.

An FFF with $d$, $n$, $\ell$ as in the definition has \textit{training size} of $(2^d - 1)n + 2^d\ell$ -- these are all the neurons of the network, and they are all affected by optimization.
It further has \textit{inference size} of $dn + \ell$, as these are the neurons engaged to produce inference output by $\textsc{Forward}_I$.

However, only the neurons of leaves produce output, with the node neurons being involved solely in the computation of the mixture of the outputs of individual leaves.
Therefore, we say that the FFF has \textit{training width} of $2^d \ell$ and inference width $\ell$.
Note that the FFF with all weights of node networks set to $0$ is equivalent to a vanilla feedforward network with $2^d\ell$ neurons (up to a uniform rescaling of the output weights, which is learnable).
We refer to the difference between the training/inference size and width as \textit{overhead}.

\section{Experiments}
\label{section:experiments}
We conduct a number of experiments to (1) explore the effects of assigning neurons with learnable regions of influence, (2) compare the predictive performance and speed of FFFs to that of MoEs, and (3) assess the feasibility of FFFs as parts of deeper architectures.
The task of each experiment is image classification, and we evaluate the classification accuracy of the softmax of output logits in the usual way.
For FFFs, we measure the accuracy of making ``hard'' decisions at every node (i.e. we use $\textsc{Forward}_I$).

For each dataset considered, we use the designated training and test sets as provided.
We further split the full training set $9:1$ into training and validation subsets.

To compare the qualities of individual models, we measure four quantities.

\textbf{Memorization accuracy ($M_A$).}
Interpreting (fast) feedforward networks as model memories \citep{bau2020rewriting}, we measure their ability to learn the training set by computing the accuracy of overfitted networks on the training set.
That is, we train networks until their accuracy in training stops improving, and then run a test on the training data.
A resulting $M_A$ of 100\% means that the network has successfully memorized all the predictions for the training data.

\textbf{Generalization accuracy ($G_A$).}
Treating (fast) feedforward networks as predictive models in their own right, we measure their ability to correctly predict the classes of previously unseen samples in the test set.
For this, we train networks until their validation accuracy stops improving, and use the best model in terms of the validation accuracy for evaluation.

\textbf{Inference time and speedup.}
Our own implementation of the FFF algorithms is provided through \texttt{pip install fastfeedforward} and on GitHub\footref{footnote:code_link}.
We compile our implementation of $\textsc{Forward}_I$ for NVIDIA A100 GPUs using PyTorch 2.0.1 model compilation in the \texttt{reduce-overhead} mode.
We then run each model $10^4$ times with batch size 2048 on a single NVIDIA A100 GPU.
Where relevant for comparison, we report the mean \textit{inference time} $\overline{t_{\bullet}}$ per single forward pass under repeated trials, together with its standard deviation.
We further report \textit{speedup} -- the fraction $\overline{t_{FF}}/\overline{t_{FFF}}$, where $\overline{t_{FFF}}$ is the mean inference time for the given FFF model and $\overline{t_{FF}}$ is the mean inference time for the vanilla feedforward network of the same training width.
Simply put, speedup says how much faster the FFF was than the FF with the same number of neurons available for making predictions in training, measured using this choice of software and hardware.
The means and deviations for speedups are in the appendix.

\section{Explorative evaluation}
\label{section:explorative_evaluation}
To examine the nature of fast feedforward networks as an alternative to feedforward networks, we measure the effect of their parameters on their predictive performance and speed in the context.

\subsection{Evaluation with training counterparts}
\label{section:expl_training_counterparts}

\begin{table*}[t!]
\centering
\scalebox{1.00}{
    \begin{tabular}{r|l|ccc|ccc|ccc|ccc}
    
    \toprule
       \multicolumn{2}{c|}{Model} & \multicolumn{12}{c}{USPS}\\
    \midrule
    
    \multicolumn{1}{c}{} & \multicolumn{1}{c}{}  & \multicolumn{3}{c}{16} & \multicolumn{3}{c}{32} & \multicolumn{3}{c}{64} & \multicolumn{3}{c}{128} \\
    \cmidrule(r){3-5}
    \cmidrule(r){6-8}
    \cmidrule(r){9-11}
    \cmidrule(r){12-14}

    \multicolumn{1}{c}{} &
    \multicolumn{1}{c}{} &
    \multirow{1}{*}{$M_A$} &
    \multirow{1}{*}{$G_A$} &
    \multirow{1}{*}{speedup} &
   \multirow{1}{*}{$M_A$} &
    \multirow{1}{*}{$G_A$} &
    \multirow{1}{*}{speedup} &
    \multirow{1}{*}{$M_A$} &
    \multirow{1}{*}{$G_A$} &
    \multirow{1}{*}{speedup} &
    \multirow{1}{*}{$M_A$} &
    \multirow{1}{*}{$G_A$} &
    \multirow{1}{*}{speedup} \\
    
    \midrule
        \multicolumn{2}{c|}{vanilla FF} & 100.0 & 93.1 & 1.00x & 100.0 & 93.7 & 1.00x & 100.0 & 94.1 & 1.00x & 100.0 & 94.2 & 1.00x \\
    \midrule
        \multirow{3}*{\rotatebox[origin=c]{90}{FFF}}

        & $\ell=8$ & 99.3 & 92.2 & 1.07x & 99.2 & 91.8 & 1.16x & 99.2 & 92.3 & 1.53x & 99.5 & 92.1 & \bw{2.56x} \\
        & $\ell=4$ & 94.1 & 87.6 & 0.98x & 97.2 & 89.5 & 1.08x & 97.6 & 90.6 & 1.56x & 97.1 & 90.3 & \bo{2.40x} \\
        & $\ell=2$ & 92.0 & 85.5 & 0.90x & 93.4 & 86.4 & 1.07x & 90.6 & 84.4 & 1.39x & 94.3 & 88.1 & \bo{2.22x} \\
        & $\ell=1$ & 83.4 & 77.0 & 0.85x & 77.3 & 74.2 & 0.99x & 79.2 & 77.1 & 1.34x & 81.4 & 77.8 & \bo{2.12x} \\
    \midrule 
    
       \multicolumn{2}{c|}{Model} & \multicolumn{12}{c}{MNIST}\\
       
    \midrule
        \multicolumn{2}{c|}{vanilla FF} & 98.0 & 95.2 & 1.00x & 100.0 & 96.6 & 1.00x & 100.0 & 97.7 & 1.00x & 100.0 & 98.1 & 1.00x \\
    \midrule
        \multirow{3}*{\rotatebox[origin=c]{90}{FFF}}

        & $\ell=8$ & 94.6 & 93.1 & 1.13x & 96.5 & 93.9 & 1.50x & 97.7 & 94.2 & 2.20x & 99.3 & 94.9 & \bo{3.39x} \\
        & $\ell=4$ & 91.6 & 90.8 & 1.33x & 96.2 & 93.1 & 1.35x & 96.7 & 93.3 & 2.33x & 97.6 & 93.6 & \bo{3.29x} \\
        & $\ell=2$ & 92.1 & 90.3 & 1.19x & 94.0 & 91.4 & 1.48x & 95.2 & 92.1 & 2.33x & 96.2 & 92.4 & \bo{3.47x} \\
        & $\ell=1$ & 91.7 & 89.9 & 1.04x & 94.4 & 92.0 & 1.26x & 94.5 & 91.4 & 1.91x & 94.1 & 92.0 & \bw{3.93x} \\

    \midrule 
    
       \multicolumn{2}{c|}{Model} & \multicolumn{12}{c}{FashionMNIST}\\
       
    \midrule
        \multicolumn{2}{c|}{vanilla FF} & 91.0 & 86.4 & 1.00x & 94.8 & 87.8 & 1.00x & 98.5 & 89.0 & 1.00x & 99.3 & 89.6 & 1.00x \\
    \midrule
        \multirow{3}*{\rotatebox[origin=c]{90}{FFF}}

        & $\ell=8$ & 86.7 & 84.2 & 1.34x & 87.8 & 85.2 & 1.44x & 88.8 & 85.2 & 2.02x & 90.5 & 86.1 & \bw{3.78x} \\
        & $\ell=4$ & 86.4 & 83.3 & 1.27x & 86.6 & 84.5 & 1.32x & 89.1 & 85.1 & 2.02x & 89.0 & 85.4 & \bo{3.41x} \\
        & $\ell=2$ & 84.5 & 83.0 & 1.24x & 85.4 & 82.9 & 1.34x & 87.2 & 84.1 & 2.01x & 87.3 & 84.3 & \bo{3.28x} \\
        & $\ell=1$ & 79.7 & 78.4 & 1.04x & 79.4 & 77.8 & 1.29x & 79.9 & 79.5 & 1.90x & 78.7 & 77.7 & \bo{2.92x} \\
    \bottomrule
    \end{tabular}
}

\caption{
    The results of the explorative experimentation on FFFs.
    Reading top-to-bottom shows the effect of decreasing the leaf size and correspondingly increasing the depth.
    \textit{Left-to-right:} The effect of increasing the training width and model depth while keeping the leaf size constant.
    \textit{Diagonally bottom-left-to-top-right:} The effect of keeping the depth constant while increasing the leaf size and training width.
    \bo{Emphasis} and \bw{emphasis} mark the best speedups per $\ell$, dataset and dataset, respectively. 
}
\label{table:small_results}
\end{table*}

\textbf{Subject.} We investigate the relationship between the configuration of leaf size, depth, and training width and the memorization and generalization performance of fast feedforward networks.
For each FFF, we also consider the performance of a FF of the same training width.
We make the comparison with them having the same training width rather than training size since only leaf neurons are directly involved in computing the classification prediction for the inputs given.

\paragraph{Method.}
We train fast feedforward networks for training widths $w=16,32,64,128$, leaf sizes $\ell = 1,2,4$, datasets USPS \cite{hull1994database}, MNIST \cite{lecun2010mnist}, and FashionMNIST \cite{xiao2017fashion}.
For each $w,\ell$ configuration we compute the depth as $\log_2 (w/\ell)$.
The set of widths has been chosen on purpose: notice that any fast feedforward network with $w, \ell$ as above has inference width smaller than $16$ -- the narrowest of our configurations.
For each width, we further train a vanilla feedforward network as a baseline.

We feed the networks flattened images.
For ease of comparison, we use batch size 256 and pure SGD optimization with learning rate of $0.2$ irrespective of the size or the depth of the networks, but we note that deeper FFFs have benefited from larger batch sizes and smaller learning rates.
We engage the hardening loss with scale parameter $h=3.0$.
We execute 10 runs for each configuration, and since this is an evaluation of architectural limits, we report the performance of the best model.
Means and deviations are in the appendix.

\paragraph{Discussion.}
Our results are listed in Table~\ref{table:small_results}.
A visualization of the hardening process can be found in Figure~\ref{figure:fff_entropy_evolution} of the appendix.
The general observations are that each of: increasing width, increasing leaf size, and increasing leaf size while keeping the depth constant; universally help memorization and generalization performance.
We make several specific observations in relation to our contributions.

\textbf{FFFs perform comparably to FFs.}
For sufficiently large widths and depths on USPS and MNIST, fast feedforward networks are only slightly (2-3\%) worse than vanilla feedforward networks.
Coincidentally, these are also the configurations in which FFFs deliver the best inference speed improvements over classical feedforward layers as measured on our hardware.

Notice the performances of the FFFs with $w=128, \ell=8$ across datasets relative to FFs with $w=16$.
The performance is remarkably close and even exceeds that of FFs, all that while the inference size of these FFFs ($12$) remains below that of the FFs ($16$).

On FashionMNIST we observe the same trends but note that an FFF beyond our testing range ($w=512, \ell=8$) was eventually necessary to bring FFFs close ($M_A$=97.1, $G_A$=88.1) to the performance of FFs.

\begin{figure*}[t!]
    \centering
    \includegraphics[width=0.95\linewidth]{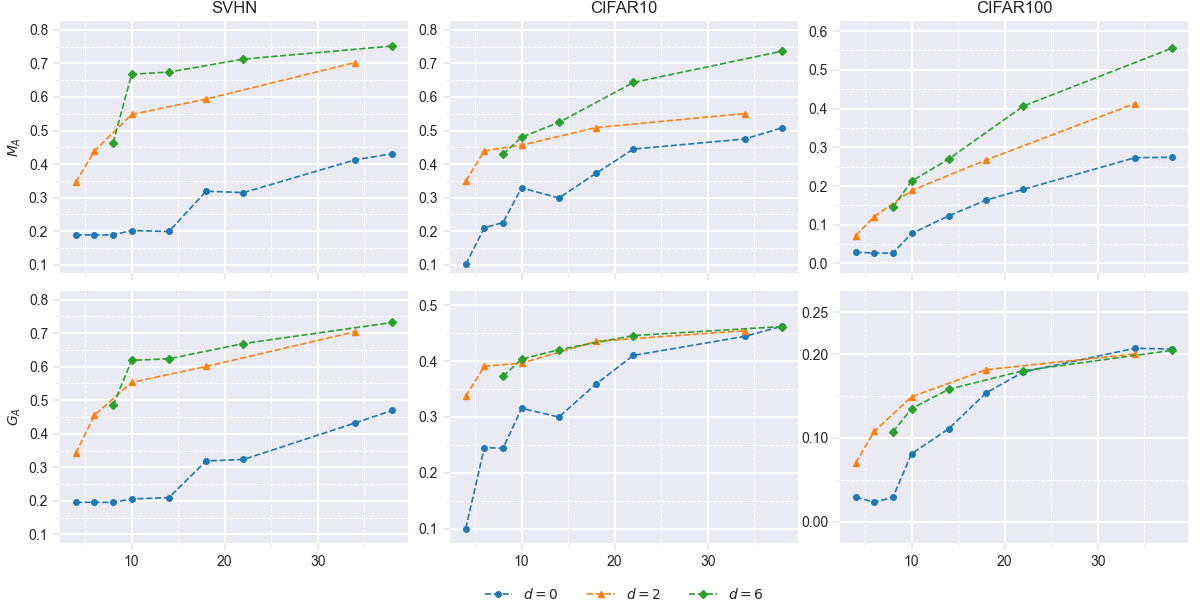}
    \caption{
        A visualization of the comparison of memorization and generalization performance of fast feedforward ($d$=2,6) and feedforward ($d$=0) networks.
        \textit{Horizontally:} the inference size in neurons.
        \textit{Vertically:} accuracy.
    }
    \label{figure:inference_counterparts}
\end{figure*}

\textbf{Speedup increases with width.}
The wider and deeper networks become while keeping the leaf size constant, the more significant the inference speed improvement delivered by the fast feedforward architecture.

\textbf{Constant width leads to a speed-performance trade-off.}
If the training width is kept fixed, there is clearly a trade-off between increasing depth (and therefore increasing speed) and performance (cf. Table~\ref{table:small_results} top-to-bottom).

Increasing the depth of the network while keeping the training width constant results in a gradual decrease in performance due to the FFF having smaller leaves, hence our note in the ``User manual'' (cf. Introduction).
Later we will see that in deeper architectures the trade-off in constant-width networks between inference speed and performance decrease due to small leaf size appears to be lessened in multi-layer architectures.

\textbf{Large-depth small-leaf networks exhibit overfragmentation.}
We observe (USPS, FashionMNIST) that decreasing the leaf size while increasing depth (keeping the width constant) leads to quickly worsening memorization and generalization performance.
The difference is particularly stark with greater depths ($\ell$=1,2, $w$=64, 128).
This is well explained by overfragmentation, since especially for $\ell=1,w=128$, each leaf receives only $2$ samples per batch on average.
The results on USPS for $\ell$=1, $w$=16,32 are a model example of this: we see that FFF with $\ell$=1,$d$=4 delivers $M_A$ of $83.4$, but that its deeper cousin $\ell$=1,$d$=5 -- which has \textit{more} of the same-sized leaves at its disposal -- yields $M_A$ of only $77.3$.

\paragraph{TL;DR.} FFFs give predictive performance comparable to FFs of the same training width, are faster as the training width increases, and if pushed to the limit exhibit speed-performance trade-offs and overfragmentation.

\subsubsection{Evaluation with inference counterparts}
\label{section:expl_inference_counterparts}

\paragraph{Subject.}
Similarly to the experimentation above with the main consideration for training width, we now evaluate fast feedforward networks of varying depths and leaf sizes with respect to their inference size.
We make the inference size the point of comparison with feedforward networks since, unlike the inference width, it is directly proportional to the computational cost of inference.

\paragraph{Method.}
We train fast feedforward networks for $\ell=2,4,6,8,16,32$, $d = 2,6$, datasets SVHN \citep{svhn}, CIFAR10, CIFAR100 \citep{cifar10}, and for each $\ell,d$ configuration we compute the inference size as $\ell + d$.
For each inference size, we further train a vanilla feedforward network as a baseline.

We train the networks with all parameters as above except $h$, where we do not engage the hardening loss ($h=0$) as we found that the hardening tended to occur on its own.
We execute 10 runs for each configuration and report the best performances.

\begin{table*}[t!]
\centering
\scalebox{1.00}{
    \begin{tabular}{l|cccc|cccc|cccc}
    
    \toprule
       \multicolumn{1}{c|}{Width} & \multicolumn{12}{c}{Model}\\
    \midrule
    
    \multicolumn{1}{c}{}  & \multicolumn{4}{c}{feedforward} & \multicolumn{4}{c}{mixture-of-experts ($e$=16, $k$=2)} & \multicolumn{4}{c}{fast feedforward ($\ell$=32)} \\
    \cmidrule(r){2-5}
    \cmidrule(r){6-9}
    \cmidrule(r){10-13}

    \multicolumn{1}{c}{} &
    \multirow{1}{*}{$M_A$} &
    \multirow{1}{*}{ETT} &
    \multirow{1}{*}{$G_A$} &
    \multirow{1}{*}{ETT} &
    \multirow{1}{*}{$M_A$} &
    \multirow{1}{*}{ETT} &
    \multirow{1}{*}{$G_A$} &
    \multirow{1}{*}{ETT} &
    \multirow{1}{*}{$M_A$} &
    \multirow{1}{*}{ETT} &
    \multirow{1}{*}{$G_A$} &
    \multirow{1}{*}{ETT} \\
    
    \midrule
        $w=64$ & 87.2 & 307 & 49.3 & 55 
            & 57.8 & 5354 & 29.4 & 4880 
            & 85.8 & 302 & 45.9 & 22 \\
        $w=128$ & 95.5 & 200 & 51.5 & 46 
            & 62.0 & 6074 & 33.6 & 938 
            & 90.1 & 305 & 45.5 & 22 \\
        $w=256$ & 99.9 & 105 & 52.0 & 48 
            & 62.4 & 2001 & 33.9 & 372
            & 91.2 & 244 & 44.4 & 17 \\
        $w=512$ & 99.9 & 85 & 52.4 & 31 
            & 65.4 & 3834 & 34.5 & 315 
            & 96.2 & 175 & 43.7 & 10 \\
        $w=1024$ & 99.9 & 82 & 53.0 & 21 
            & 65.3 & 1575 & 35.2 & 327 
            & 96.0 & 180 & 41.3 & 9 \\
    \bottomrule
    \end{tabular}
}

\caption{
    The results of the comparison of feedforward, mixture-of-experts, and fast feedforward networks, for various training widths.
    The inference width is fixed to 32 for mixture-of-experts and fast feedforward networks.
    The ETT columns to the right of metric columns list the ``epochs to train'', i.e. the number of training epochs that have elapsed until the score to the left was observed.
}
\label{table:comparative_results}
\end{table*}

\paragraph{Discussion.}
Our results are sparse because of all the different possible sums of leaf size and depth and are shown in Figure~\ref{figure:inference_counterparts}.
Aligned with intuition, we observe that bigger depth and larger leaf sizes lead to better memorization and generalization performance.

Further, \textbf{FFFs outperform FFs of the same inference size.}
FFFs of varying depths and sizes consistently outperform the FFs with widths equal to the FFF inference sizes, both in terms of $M_A$ and $G_A$.
In terms of $M_A$, the difference is stark and grows with the depth and leaf size.
In terms of $G_A$, FFFs initially gain an edge over FFs, but later the performances of all models unite toward plateauing out at the limit of the naive, single-layer feedforward-only approach.

\paragraph{TL;DR.} FFFs deliver performance more readily than FFs of the same inference width.

\section{Comparative evaluation}
\label{section:comparative_evaluation}
The direct contender architecture to fast feedforward, coming along with its own set of design parameters, is the mixture-of-expert layer, which we take in its original form \cite{shazeer2017outrageously}.

\subsection{Predictive performance comparison}

\paragraph{Subject.}
We compare FFFs against MoEs and FFs of varying training widths in terms of their predictive performance.
We keep the leaf and expert width constant and focus on the ability of the architectures to deliver good memorization and generalization properties as well as on the training compute necessary to reach those properties.

\paragraph{Method.}
We experiment on the unaugmented CIFAR10 dataset.
We train feedforward, mixture-of-experts, and fast feedforward networks of increasing size so that they always agree in the training width.
To keep the inference width the same, we set the leaf width to $32$ and expert width to $16$ with always engaging $k=2$ experts.
Note that while single-expert networks can be used for inference, they are not able to propagate gradients to the gating network (cf. \citet{shazeer2017outrageously}).
We take widths $w=2^6,2^7,\dots,2^{10}$, which correspond to 16- to 64-expert MoE networks and FFFs of depths 1 to 5.
To encourage importance equality and load balancing in MoEs, we set $w_{importance}=w_{load}=0.1$ in line with previous work.
To encourage FFF hardening, we use $h=3.0$.
We train all models width batch size 4096 for 7000 epochs at most, with early stopping after 350 epochs where no improvement in the respective validation accuracies is seen.
All models have been trained with the Adam optimizer, learning rate of $0.001$, with the learning rate halving on 250-epoch training accuracy plateaus.

\paragraph{Discussion.}
The results are listed in Table~\ref{table:comparative_results}.
On the outset, we observe that the $M_A$ and $G_A$ benefit from larger training width across all models except $G_A$ on FFFs, where we see the unmitigated localized overfitting negatively affecting the performance.

\textbf{FFFs are the fastest to deliver $M_A$ and $G_A$.}
We see that the FFFs are the fastest (in terms of ETT) to deliver both $M_A$ and $G_A$, but, consistently with our previous explorative evaluation, deliver slightly lower $M_A$ than FFs of the same training width, and suffer from localized overfitting with the increasing depth.

\textbf{FFFs outperform MoEs of the same training width.}
We observe that FFFs consistently deliver better $M_A$ and $G_A$ scores than the MoE networks of the same training width.
We further see that they do so at ETTs smaller by an order of magnitude.
We attribute this difference mainly to the learnably controlled noise introduced to the expert mixture computation to aid load balancing and generalization.
Without the noise, however, MoE networks would overfit to learn only with a handful of experts.
We also experimented with varying values of $w_{\text{importance}}$ and $w_{\text{load}}$, but we found those to be broadly detrimental to the load balancing effort. Our final values of batch importance and load were consistent with those arrived at in \citet{shazeer2017outrageously}.

\paragraph{TL;DR.}
FFFs deliver representational power more readily than the MoEs of equal training widths.

\subsection{Inference speed comparison}

\paragraph{Subject.}
Since the operations involved in the computation of the expert/leaf network output are the same, the difference in inference speed between mixture-of-experts and fast feedforward networks comes solely from the functioning of the gating/lookup mechanism.
We therefore keep the expert/leaf width constant and measure the time needed to execute inference forward passes of feedforward, mixture-of-experts, and fast feedforward networks across repeated trials for increasing numbers of experts/leaves (i.e. wider and wider networks).

To add realism, we simulate the conditions of a BERT-base \cite{devlin2018bert} feedforward layer, setting the input and output widths of all neurons to $768$ each.

\begin{figure}[h]
    \centering
    \includegraphics[width=\linewidth]{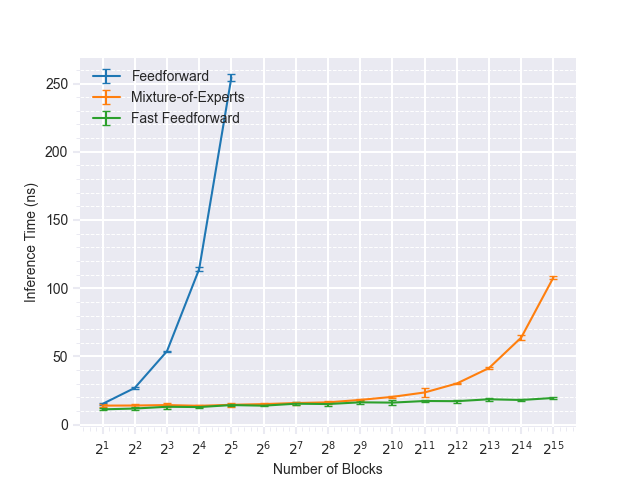}
    \caption{
        A visualization of the performance measurement results.
        The horizontal axis denotes the number of blocks/experts/leaves and is scaled logarithmically, the point values are the mean inference times per single forward pass, and the error bars show the standard deviation.
    }
    \label{figure:ffvsmoevsfff}
\end{figure}

\begin{figure}[h]
    \centering
    \includegraphics[width=\linewidth]{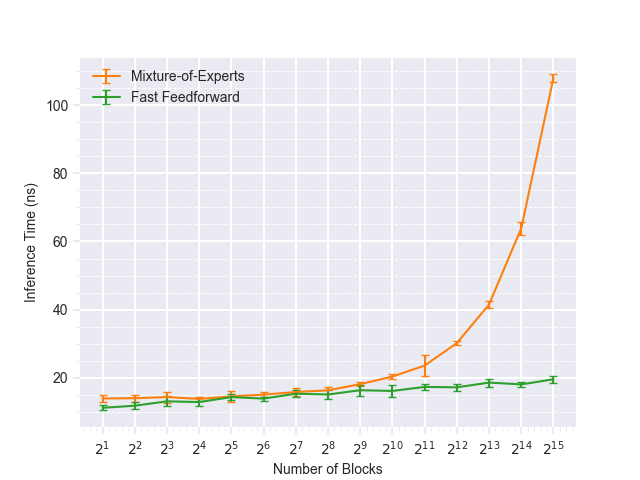}
    \caption{
        A close-up on the visualization of the performance measurement results.
        The axes and values are as in Figure~\ref{figure:ffvsmoevsfff}.
    }
    \label{figure:moevsfff}
\end{figure}

\paragraph{Method.}
We consider FF models of width $32 \times 2^1$ to $32 \times 2^5$, where we highlight the 32 neuron blocks for direct comparison with the other models.
We further evaluate MoE models with expert width $e=32$ and $2^1$ to $2^{15}$ experts, and FFF models with leaf width $e=32$ and depths $1$ to $15$.
To eliminate the effect of the mixture computation on our measurements, we keep $e=\ell$ and set $k=1$, even though this MoE parameter configuration is not trainable (cf. above).
When measuring, each model performs inference on BERT-base inputs with batch size 256 exactly 20000 times.

\paragraph{Discussion.}
The inference speed measurement results are visualized in Figures~\ref{figure:ffvsmoevsfff}--\ref{figure:moevsfff}.
We observe that both MoEs and FFFs offer significant acceleration to the inference speed when compared to the FF baseline.
However, Figure~\ref{figure:moevsfff}, shows the clear tendency of MoE model inference time to grow exponentially with the exponent of the expert count, in stark contrast with the linear relationship between the two exhibited by the FFF models.
This is fully aligned with our theoretical analysis of the inference time complexity of the two architectures.

\paragraph{TL;DR.}
We have experimentally confirmed the exponential difference between the inference time complexity of the MoE's and FFF's internal mechanism.

\section{Fast feedforward layers as building blocks}
\label{section:building_blocks}

\begin{table*}[t!]
\centering
\scalebox{1.00}{
    \begin{tabular}{c|l|rrrrr|c|c}
    
    \toprule
       \multicolumn{2}{c|}{Model} & \multicolumn{7}{c}{Property}\\
    \midrule

    \multicolumn{2}{c|}{} &
    
    \multirow{1}{*}{depth} &
    \multirow{1}{*}{training width} &
    \multirow{1}{*}{training size} &
    \multirow{1}{*}{inference width} &
    \multirow{1}{*}{inference size} &
    \multirow{1}{*}{speedup} &
    \multirow{1}{*}{$G_A$} \\
    
    \midrule
        FF
        
        & $w=128$    & --&	128&  128 (100\%) &  128 (100\%)& 128 (100\%) & 1.00x & 84.7 \\
        
    \midrule
        \multirow{6}*{\rotatebox[origin=c]{90}{fast FF}}
        
        & $\ell=32$ & 2 &	128&  131 (102\%)&  32\hspace{5pt} (25\%)&	34\hspace{5pt} (27\%)&	2.44x & 83.6  \\
        & $\ell=16$ & 3 &	128&  135 (105\%)&  16\hspace{5pt} (13\%)&	19\hspace{5pt} (15\%)&	2.80x & 83.2  \\
        & $\ell=8$  & 4 &	128&  143 (112\%)&  8\hspace{10pt}  (6\%)&	12\hspace{10pt} (9\%)&	3.29x & 82.8  \\
        & $\ell=4$  & 5 &	128&  159 (124\%)&  4\hspace{10pt}  (3\%)&	9\hspace{10pt}  (7\%)&	3.39x & 81.6  \\
        & $\ell=2$  & 6 &	128&  191 (149\%)&  2\hspace{10pt}  (1\%)&	8\hspace{10pt}  (6\%)&	3.47x & 80.1  \\
        & $\ell=1$  & 7 &	128&  255 (199\%)&  1\hspace{10pt}  (1\%)&	8\hspace{10pt}  (6\%)&	3.93x & 79.8  \\
    \bottomrule 
    
    \end{tabular}
}

\caption{
    The results of the testing of vision transformers leveraging feedforward and fast feedforward layers.
    All sizes are given in neurons. Bracketed percentages describe quantities relative to their counterparts in the vanilla feedforward layers.
    $G_A$ is the generalization accuracy of the fully trained vision transformer and ``speedup'' gives the performance improvement over vanilla feedforward layers in our testing setup.
}
\label{table:vit_results}
\end{table*}

\paragraph{Subject.}
We demonstrate that fast feedforward networks can be used as layers in place of standard feedforward layers in the transformer architecture, thus giving it a significant performance boost.
Previously, we noted that leaf sizes that are too small for a given problem may lead to the occurrence of overfragmentation.
Here we push our experimental setup to the limit in terms of leaf size and investigate the effect of overfragmentation in a deep transformer.

\paragraph{Method.}
We experiment on the CIFAR10 dataset of 32x32-pixel 3-channel images, with random horizontal, vertical flipping, and random linear augmentations (translate, rotate, scale).
As models, we use 4-layer vision transformers with patch size 4, hidden dimension 128, input dropout $0.1$, and no layer dropout.

We consider vision transformers with their feedforward layers replaced by fast feedforward layers of training width $128$, and a baseline vision transformer with feedforward layers of width $w=128$.
The fast feedforward layers have leaf size $\ell=1,2,4,8,16,32$ and depths $\log_2 (w / \ell)$.
We try three levels of hardening: $h=5, 10, \infty$, where $\infty$ denotes that the FFF tree has been effectively frozen from the beginning (i.e. the boundaries are not trainable).
We use Adam optimizer with the initial learning rate of $4e-4$ and learning rate halving on 50-epoch validation accuracy plateaus.
For each $\ell$,$d$-configuration, we report the generalization performance of the best model and the measured speedup at the feedforward layers (not the whole transformer).

\paragraph{Discussion.}
The results of our experimentation can be seen in Table~\ref{table:vit_results}.
A visualization of the hardening process across the layers of the transformer can be found in Figure~\ref{figure:vit_entropy_evolution} of the appendix.
In line with our assessment of the algorithm complexity, the measured speedup at the feedforward layers increases with decreasing leaf size.
Further:

\textbf{Single-neuron FFFs suffice.}
We observe that even fast feedforward layers with inference width $1$ are sufficient for the vision transformer to deliver reasonable performance, with relative decrease in performance of only $5.8\%$.

\textbf{The effects of overfragmentation are suppressed.}
We observe that the generalization performance of $G_A$ suffers only relatively mildly due to the increase in depth and decrease in leaf size, which is in stark contrast with the results of Table~\ref{table:small_results}. 
We attribute this to the depth of the transformer and take it as an encouraging sign of the feasibility of FFFs as replacements for FFs.

\paragraph{TL;DR.}
Fast feedforward layers can deliver inference acceleration in vision transformers over feedforward layers of the same training width at the cost of a small performance decrease.

\section{Related work}
\label{section:related_work}
Our work overlaps with the research efforts in two areas of inference acceleration.

\paragraph{Conditional execution.}
Although nowadays largely inactive due to the tendency to move away from custom architectures, modified designs for MLP and CNN models were proposed to allow for their partial execution where possible.

A number of methods were proposed \citep{davis2013low,bengio2013estimating,bengio2015conditional,almahairi2016dynamic} to learn either policy distributions or additional controlling neural components to decide which blocks of layer neurons to execute during forward pass.

In comparison, fast feedforward networks completely conceal the learning of leaf regions from the user (save from the hyperparameter $h$ if used) and come in an inference-ready form once trained, requiring no adjustment when included as a part of transformers.

In a notable generalization of this line of work to deep architectures, \citet{ioannou2016decision} proposed an approach peer to deep convolutional neural networks that learns to route the input through a sequence of chosen intermediate layers.
While our method quickly routes the signal to a single-leaf feedforward neural network, it draws no comparison to deep networks.

\paragraph{Modular ``mixture-of-experts'' models.}
Very large models practically demand modularity.
The most straightforward way to modularize large transformer models in order to reduce their inference cost is to subdivide their feedforward layers into $n$ blocks of neurons, and then train a controlling classifier to choose which block to involve in forward pass.
This is usually done by training a wide softmax-activated linear layer to produce a stochastic vector of mixture scores to be applied to the outputs per block in order to produce the final output.
Several variants of this method have been proposed and tested across a variety of large models \citep{shazeer2017outrageously,lepikhin2020gshard,fedus2022switch}.

We thoroughly compare fast feedforward to mixture-of-expert networks in earlier sections.
To briefly summarise, the mixture-of-experts approach reduces the layer inference width by a factor of $n/k$, where $k$ is the number of best-scoring blocks to engage in inference, but requires $\mathcal{O}(n)$ time to select the $k$ blocks, and often relies on controlled randomization to avoid the formation of a strong preference for only a handful experts.
This holds true even when multiple layers of expert mixers are introduced.
In direct comparison, a fast feedforward network of depth $d=\log_2 n$ reduces the inference by a factor of $n$ and requires only $\mathcal{O}(d) = \mathcal{O}(\log n)$ time to decide on which leaf to use.
Admittedly, to compensate for the effect of having only one leaf to make the decision, the leaves of the fast feedforward layer might have to be slightly wider than the blocks of the corresponding mixture-of-experts layer.

\paragraph{Regionalization.}
An additional advantage of FFF over all of the related work is that there is a direct correspondence between parts of the network used in inference and algebraically identifiable regions of the input space.
This can be leveraged to mitigate catastrophic forgetting when editing models and to significantly reduce replay data budgets by applying the learned partition of the input space to partition the training data.

\appendix

\bibliography{aaai24}

\begin{thebibliography}{17}
\providecommand{\natexlab}[1]{#1}

\bibitem[{Almahairi et~al.(2016)Almahairi, Ballas, Cooijmans, Zheng,
  Larochelle, and Courville}]{almahairi2016dynamic}
Almahairi, A.; Ballas, N.; Cooijmans, T.; Zheng, Y.; Larochelle, H.; and
  Courville, A. 2016.
\newblock Dynamic capacity networks.
\newblock In \emph{International Conference on Machine Learning}, 2549--2558.
  PMLR.

\bibitem[{Bau et~al.(2020)Bau, Liu, Wang, Zhu, and Torralba}]{bau2020rewriting}
Bau, D.; Liu, S.; Wang, T.; Zhu, J.-Y.; and Torralba, A. 2020.
\newblock Rewriting a deep generative model.
\newblock In \emph{Computer Vision--ECCV 2020: 16th European Conference,
  Glasgow, UK, August 23--28, 2020, Proceedings, Part I 16}, 351--369.
  Springer.

\bibitem[{Bengio et~al.(2015)Bengio, Bacon, Pineau, and
  Precup}]{bengio2015conditional}
Bengio, E.; Bacon, P.-L.; Pineau, J.; and Precup, D. 2015.
\newblock Conditional computation in neural networks for faster models.
\newblock \emph{arXiv preprint arXiv:1511.06297}.

\bibitem[{Bengio, L{\'e}onard, and Courville(2013)}]{bengio2013estimating}
Bengio, Y.; L{\'e}onard, N.; and Courville, A. 2013.
\newblock Estimating or propagating gradients through stochastic neurons for
  conditional computation.
\newblock \emph{arXiv preprint arXiv:1308.3432}.

\bibitem[{Bentley(1975)}]{bentley1975multidimensional}
Bentley, J.~L. 1975.
\newblock Multidimensional binary search trees used for associative searching.
\newblock \emph{Communications of the ACM}, 18(9): 509--517.

\bibitem[{Davis and Arel(2013)}]{davis2013low}
Davis, A.; and Arel, I. 2013.
\newblock Low-rank approximations for conditional feedforward computation in
  deep neural networks.
\newblock \emph{arXiv preprint arXiv:1312.4461}.

\bibitem[{Devlin et~al.(2018)Devlin, Chang, Lee, and
  Toutanova}]{devlin2018bert}
Devlin, J.; Chang, M.-W.; Lee, K.; and Toutanova, K. 2018.
\newblock Bert: Pre-training of deep bidirectional transformers for language
  understanding.
\newblock \emph{arXiv preprint arXiv:1810.04805}.

\bibitem[{Fedus, Zoph, and Shazeer(2022)}]{fedus2022switch}
Fedus, W.; Zoph, B.; and Shazeer, N. 2022.
\newblock Switch transformers: Scaling to trillion parameter models with simple
  and efficient sparsity.
\newblock \emph{The Journal of Machine Learning Research}, 23(1): 5232--5270.

\bibitem[{Hull(1994)}]{hull1994database}
Hull, J.~J. 1994.
\newblock A database for handwritten text recognition research.
\newblock \emph{IEEE Transactions on pattern analysis and machine
  intelligence}, 16(5): 550--554.

\bibitem[{Ioannou et~al.(2016)Ioannou, Robertson, Zikic, Kontschieder, Shotton,
  Brown, and Criminisi}]{ioannou2016decision}
Ioannou, Y.; Robertson, D.; Zikic, D.; Kontschieder, P.; Shotton, J.; Brown,
  M.; and Criminisi, A. 2016.
\newblock Decision forests, convolutional networks and the models in-between.
\newblock \emph{arXiv preprint arXiv:1603.01250}.

\bibitem[{Krizhevsky, Hinton et~al.(2009)}]{cifar10}
Krizhevsky, A.; Hinton, G.; et~al. 2009.
\newblock Learning multiple layers of features from tiny images.

\bibitem[{LeCun, Cortes, and Burges(2010)}]{lecun2010mnist}
LeCun, Y.; Cortes, C.; and Burges, C. 2010.
\newblock MNIST handwritten digit database.
\newblock \emph{ATT Labs [Online]. Available:
  http://yann.lecun.com/exdb/mnist}, 2.

\bibitem[{Lepikhin et~al.(2020)Lepikhin, Lee, Xu, Chen, Firat, Huang, Krikun,
  Shazeer, and Chen}]{lepikhin2020gshard}
Lepikhin, D.; Lee, H.; Xu, Y.; Chen, D.; Firat, O.; Huang, Y.; Krikun, M.;
  Shazeer, N.; and Chen, Z. 2020.
\newblock Gshard: Scaling giant models with conditional computation and
  automatic sharding.
\newblock \emph{arXiv preprint arXiv:2006.16668}.

\bibitem[{Netzer et~al.(2011)Netzer, Wang, Coates, Bissacco, Wu, and Ng}]{svhn}
Netzer, Y.; Wang, T.; Coates, A.; Bissacco, A.; Wu, B.; and Ng, A.~Y. 2011.
\newblock Reading digits in natural images with unsupervised feature learning.

\bibitem[{Shazeer et~al.(2017)Shazeer, Mirhoseini, Maziarz, Davis, Le, Hinton,
  and Dean}]{shazeer2017outrageously}
Shazeer, N.; Mirhoseini, A.; Maziarz, K.; Davis, A.; Le, Q.; Hinton, G.; and
  Dean, J. 2017.
\newblock Outrageously large neural networks: The sparsely-gated
  mixture-of-experts layer.
\newblock \emph{arXiv preprint arXiv:1701.06538}.

\bibitem[{Vaswani et~al.(2017)Vaswani, Shazeer, Parmar, Uszkoreit, Jones,
  Gomez, Kaiser, and Polosukhin}]{vaswani2017attention}
Vaswani, A.; Shazeer, N.; Parmar, N.; Uszkoreit, J.; Jones, L.; Gomez, A.~N.;
  Kaiser, {\L}.; and Polosukhin, I. 2017.
\newblock Attention is all you need.
\newblock \emph{Advances in neural information processing systems}, 30.

\bibitem[{Xiao, Rasul, and Vollgraf(2017)}]{xiao2017fashion}
Xiao, H.; Rasul, K.; and Vollgraf, R. 2017.
\newblock Fashion-mnist: a novel image dataset for benchmarking machine
  learning algorithms.
\newblock \emph{arXiv preprint arXiv:1708.07747}.

\end{thebibliography}



\clearpage

\appendix

\section{Extended results of Table~\ref{table:small_results}}
Table~\ref{table:big_results} lists the means and standard deviations of our explorative evaluation of fast feedforward networks in comparison with feedforward networks.

On top of the observations made in the main text, we see that the variance of $M_A$ and $G_A$ when performing repeated runs increases with the decreasing leaf size and is most clearly seen for small training widths.

\section{Visualization of the hardening process}
To provide some insight and intuition into the hardening of the node decision boundaries, we track the batch mean entropies across the nodes of fast feedforward layers and report on their evolution as training progresses visually.

\begin{figure}[t]
    \centering
    \includegraphics[width=\linewidth]{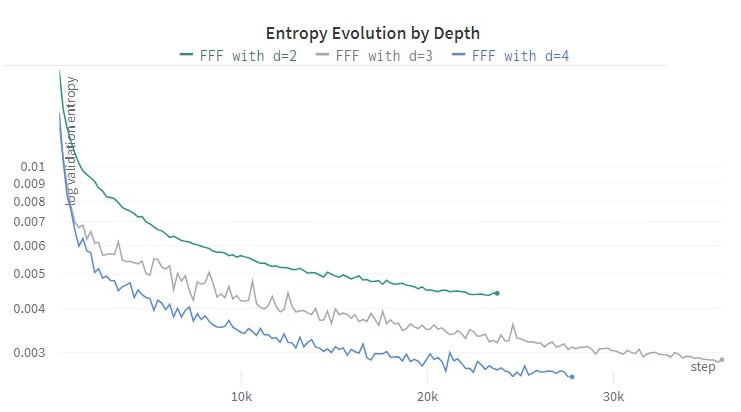}
    \caption{The evolution of batched mean decision entropy in a fast feedforward layer with $\ell=8,d=2,3,4,h=3.0$, trained on the MNIST dataset.}
    \label{figure:fff_entropy_evolution}
\end{figure}

Figure~\ref{figure:fff_entropy_evolution} shows the evolution of the batched mean entropy for three models of varying depths whose training was terminated by early stopping on validation $\textsc{Forward}_I$ accuracy plateau.
We observe that by keeping the dataset (and therefore the complexity of the input space) and leaf size constant, the batch mean accuracy of deep layers converges faster for deeper networks.
We attribute this simply to the fact that the deeper the fast feedforward network in this experiment, the more leaves were available to selectively cater to sub-spaces of the input space, allowing the networks to make cleaner separations between the regions learned.

\begin{figure}[t]
    \centering
    \includegraphics[width=\linewidth]{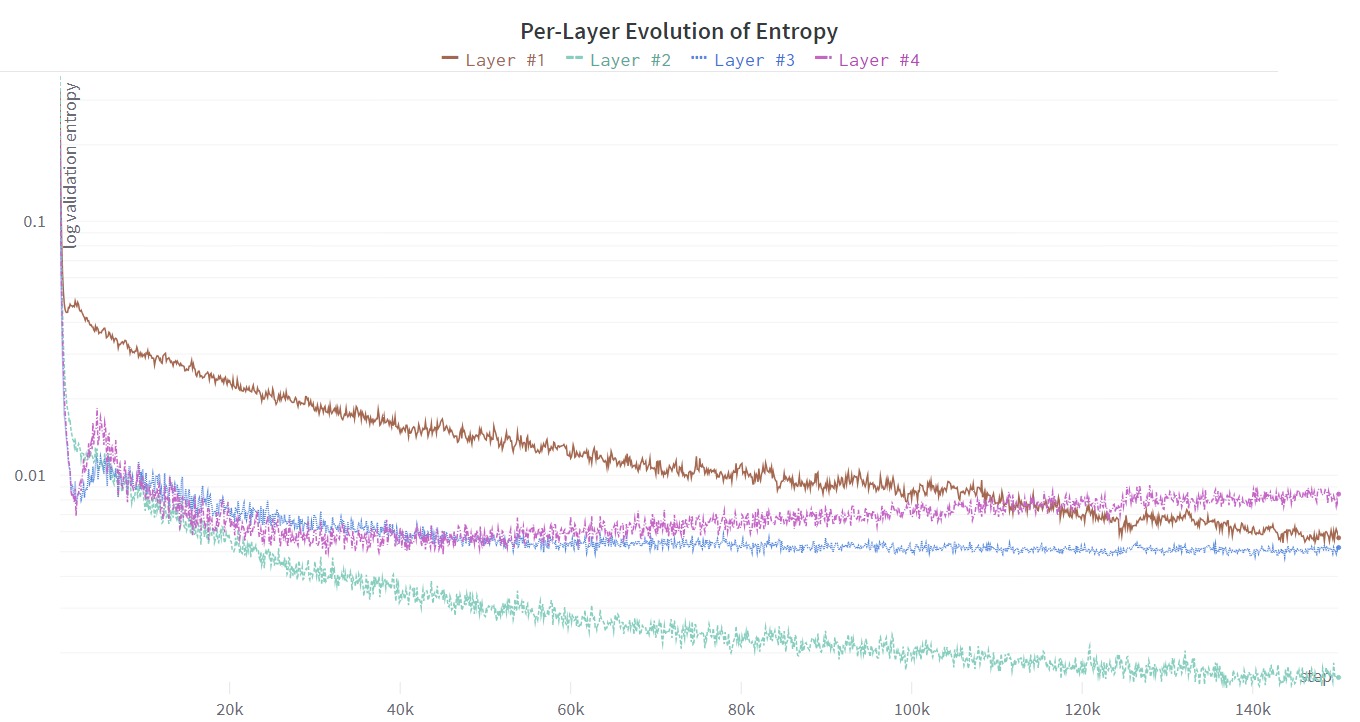}
    \caption{The evolution of batched mean decision entropies across a vision 4-layer transformer of dimension 128 equipped with fast feedforward layers of $\ell=32,d=2,h=0.10$, trained on the CIFAR10 dataset.}
    \label{figure:vit_entropy_evolution}
\end{figure}

Figure~\ref{figure:vit_entropy_evolution} reports the evolution of batched mean entropies per layer for a 4-layer visual transformer model with fast feedforward networks in place of the vanilla feedforward networks.
We notice that early into the training, entropies of the layers closer to the input (i.e. the earlier/lower layers) are faster to converge.
However, we see that the entropy of the second layer then ceases to meaningfully decrease, and that the entropies of the first layer actually begin to climb.
This can be explained by hardened decision boundaries showing signs of bottleneck behavior in higher layers of the transformer, thus necessitating that the lower layers contribute with learned representations of higher quality.

Interpreting the stalling or climbing of the entropy as a sign of the need for more representational power, we note across multiple experiments that the ideal shape for a vision transformer, even on a relatively simple dataset such as CIFAR10, appears to be more toward the shape of a pyramid than a uniform-width line of layers.

\section{Relationship to multi-gating MoE networks}
More recent literature on mixture-of-experts networks \citep{lepikhin2020gshard,fedus2022switch} suggests the use of a hierarchical, two-layer gating structure.
Save from the inherent randomness of the gating network selection, this can be seen as a small step towards FFFs.
However, FFFs attend to clearly specified regions of space rather than being noisily voted in as the most appropriate experts for a given input (which we have shown gives them much better training and inference properties), and further, even under such an approach, the inference time reduction of mixture-of-expert networks remains constant in the width of the target network.

\section{Code}
We provide our implementations as a Python package built on top of the PyTorch framework.

Just use \texttt{pip install fastfeedforward}. Documentation is provided with the code.


\newpage
\begin{sidewaystable*}
\centering
\scalebox{0.82}{
    \begin{tabular}{r|l|ccc|ccc|ccc|ccc}
    
    \toprule
       \multicolumn{2}{c|}{Model} & \multicolumn{12}{c}{USPS}\\
    \midrule
    
    \multicolumn{1}{c}{} & \multicolumn{1}{c}{}  & \multicolumn{3}{c}{16} & \multicolumn{3}{c}{32} & \multicolumn{3}{c}{64} & \multicolumn{3}{c}{128} \\
    \cmidrule(r){3-5}
    \cmidrule(r){6-8}
    \cmidrule(r){9-11}
    \cmidrule(r){12-14}

    \multicolumn{1}{c}{} &
    \multicolumn{1}{c}{} &
    \multirow{1}{*}{$M_A$} &
    \multirow{1}{*}{$G_A$} &
    \multirow{1}{*}{time} &
   \multirow{1}{*}{$M_A$} &
    \multirow{1}{*}{$G_A$} &
    \multirow{1}{*}{time} &
    \multirow{1}{*}{$M_A$} &
    \multirow{1}{*}{$G_A$} &
    \multirow{1}{*}{time} &
    \multirow{1}{*}{$M_A$} &
    \multirow{1}{*}{$G_A$} &
    \multirow{1}{*}{time} \\
    
    \midrule
        \multicolumn{2}{c|}{vanilla FF} & 100.0 ± 0.0 & 93.1 ± 0.4 & 0.13 ± 0.02ms & 100.0 ± 0.0 & 93.7 ± 0.4 & 0.16 ± 0.01ms & 100.0 ± 0.0 & 94.1 ± 0.3 & 0.24 ± 0.02ms & 100.0 ± 0.0 & 94.2 ± 0.3 & 0.41 ± 0.03ms \\
    \midrule
        \multirow{3}*{\rotatebox[origin=c]{90}{fast FF}}

        & $\ell=8$ & 99.3 ± 0.5 & 92.2 ± 0.4 & 0.12 ± 0.03ms & 99.2 ± 0.5 & 91.8 ± 0.6 & 0.14 ± 0.03ms & 99.2 ± 0.4 & 92.3 ± 0.7 & 0.16 ± 0.03ms & 99.5 ± 0.8 & 92.1 ± 0.4 & 0.16 ± 0.02ms \\
        & $\ell=4$ & 94.1 ± 0.5 & 87.6 ± 0.7 & 0.13 ± 0.03ms & 97.2 ± 3.4 & 89.5 ± 2.0 & 0.15 ± 0.03ms & 97.6 ± 2.3 & 90.6 ± 1.4 & 0.16 ± 0.02ms & 97.1 ± 1.3 & 90.3 ± 1.2 & 0.17 ± 0.02ms \\
        & $\ell=2$ & 92.0 ± 9.2 & 85.5 ± 8.0 & 0.14 ± 0.03ms & 93.4 ± 10.0 & 86.4 ± 8.7 & 0.15 ± 0.02ms & 90.6 ± 7.7 & 84.4 ± 6.0 & 0.18 ± 0.03ms & 94.3 ± 8.6 & 88.1 ± 7.1 & 0.18 ± 0.03ms \\
        & $\ell=1$ & 83.4 ± 12.1 & 77.0 ± 11.1 & 0.15 ± 0.02ms & 77.3 ± 5.8 & 74.2 ± 5.0 & 0.16 ± 0.01ms & 79.2 ± 8.1 & 77.1 ± 7.1 & 0.18 ± 0.02ms & 81.4 ± 9.2 & 77.8 ± 8.3 & 0.19 ± 0.02ms \\
    \midrule 
    
       \multicolumn{2}{c|}{Model} & \multicolumn{12}{c}{MNIST}\\
       
    \midrule
        \multicolumn{2}{c|}{vanilla FF} & 98.0 ± 0.9 & 95.2 ± 0.5 & 0.34 ± 0.11ms & 100.0 ± 0.0 & 96.6 ± 0.2 & 0.42 ± 0.06ms & 100.0 ± 0.0 & 97.7 ± 0.2 & 0.69 ± 0.10ms & 100.0 ± 0.0 & 98.1 ± 0.1 & 1.13 ± 0.06ms \\
    \midrule
        \multirow{3}*{\rotatebox[origin=c]{90}{fast FF}}

        & $\ell=8$ & 94.6 ± 19.5 & 93.1 ± 16.6 & 0.30 ± 0.11ms & 96.5 ± 2.3 & 93.9 ± 1.2 & 0.28 ± 0.05ms & 97.7 ± 4.3 & 94.2 ± 2.4 & 0.31 ± 0.06ms & 99.3 ± 1.0 & 94.9 ± 0.6 & 0.33 ± 0.08ms \\
        & $\ell=4$ & 91.6 ± 29.3 & 90.8 ± 27.2 & 0.26 ± 0.07ms & 96.2 ± 24.3 & 93.1 ± 23.9 & 0.31 ± 0.09ms & 96.7 ± 1.0 & 93.3 ± 0.6 & 0.30 ± 0.06ms & 97.6 ± 0.6 & 93.6 ± 0.5 & 0.34 ± 0.08ms \\
        & $\ell=2$ & 92.1 ± 7.3 & 90.3 ± 5.6 & 0.28 ± 0.08ms & 94.0 ± 1.4 & 91.4 ± 1.0 & 0.28 ± 0.05ms & 95.2 ± 1.8 & 92.1 ± 1.2 & 0.30 ± 0.07ms & 96.2 ± 1.4 & 92.4 ± 0.6 & 0.32 ± 0.06ms \\
        & $\ell=1$ & 91.7 ± 7.4 & 89.9 ± 6.4 & 0.33 ± 0.11ms & 94.4 ± 3.5 & 92.0 ± 3.1 & 0.33 ± 0.07ms & 94.5 ± 1.8 & 91.4 ± 1.1 & 0.36 ± 0.09ms & 94.1 ± 0.9 & 92.0 ± 0.7 & 0.29 ± 0.03ms \\

    \midrule 
    
       \multicolumn{2}{c|}{Model} & \multicolumn{12}{c}{FashionMNIST}\\
       
    \midrule
        \multicolumn{2}{c|}{vanilla FF} & 91.0 ± 0.7 & 86.4 ± 0.4 & 0.34 ± 0.10ms & 94.8 ± 0.9 & 87.8 ± 0.2 & 0.42 ± 0.07ms & 98.5 ± 0.8 & 89.0 ± 0.4 & 0.64 ± 0.07ms & 99.3 ± 0.4 & 89.6 ± 0.2 & 1.13 ± 0.05ms \\
    \midrule
        \multirow{3}*{\rotatebox[origin=c]{90}{fast FF}}

        & $\ell=8$ & 86.7 ± 12.1 & 84.2 ± 10.9 & 0.26 ± 0.07ms & 87.8 ± 17.6 & 85.2 ± 16.1 & 0.29 ± 0.10ms & 88.8 ± 5.6 & 85.2 ± 3.8 & 0.32 ± 0.05ms & 90.5 ± 1.7 & 86.1 ± 1.0 & 0.30 ± 0.06ms \\
        & $\ell=4$ & 84.5 ± 25.0 & 83.0 ± 24.5 & 0.27 ± 0.11ms & 86.6 ± 8.6 & 84.5 ± 6.4 & 0.32 ± 0.08ms & 89.1 ± 3.5 & 85.1 ± 2.3 & 0.32 ± 0.07ms & 89.0 ± 0.7 & 85.4 ± 0.7 & 0.33 ± 0.07ms \\
        & $\ell=2$ & 83.6 ± 21.0 & 82.5 ± 11.0 & 0.28 ± 0.10ms & 85.4 ± 8.4 & 82.9 ± 6.5 & 0.32 ± 0.09ms & 87.2 ± 7.1 & 84.1 ± 5.9 & 0.32 ± 0.06ms & 85.3 ± 5.2 & 81.5 ± 3.7 & 0.35 ± 0.08ms \\
        & $\ell=1$ & 86.4 ± 9.0 & 83.3 ± 8.0 & 0.33 ± 0.09ms & 79.4 ± 6.2 & 77.8 ± 5.5 & 0.33 ± 0.07ms & 79.9 ± 3.5 & 79.5 ± 3.7 & 0.34 ± 0.08ms & 78.7 ± 4.6 & 77.7 ± 3.8 & 0.39 ± 0.08ms \\
    \bottomrule
    \end{tabular}
}

\caption{
    The detailed results of the explorative experimentation on FFFs.
    Reading top-to-bottom shows the effect of decreasing the leaf size and correspondingly increasing the depth.
    \textit{Left-to-right:} The effect of increasing the training width and model depth while keeping the leaf size constant.
    \textit{Diagonally bottom-left-to-top-right:} The effect of keeping the depth constant while increasing the leaf size and training width.
}
\label{table:big_results}
\end{sidewaystable*}

\end{document}